\documentclass[runningheads]{llncs}

\usepackage[T1]{fontenc}
\usepackage{amsmath}
\usepackage{amssymb}

\usepackage{cite}
\usepackage{float}
\usepackage{subcaption}
\usepackage{multirow}
\usepackage{booktabs}
\usepackage{colortbl}
\usepackage{graphicx}
\usepackage[utf8]{inputenc}
\usepackage{kotex}
\usepackage{bbm}
\usepackage{bm}
\usepackage{array} % for extended column definitions, e.g. >{\em}c
\usepackage{blindtext}
\usepackage{comment}
\usepackage[capitalise]{cleveref}
\usepackage[table]{xcolor}
\usepackage[most,skins,raster]{tcolorbox}
\newcommand{\framework}{\textsc{EPLKG}}
\newcommand{\tabstyle}[1]{
  \setlength{\tabcolsep}{#1}
  \renewcommand{\arraystretch}{\tableCellHeight}
  \centering
  \small
}
\newcommand{\tableCellHeight}{1}
\definecolor{tabhighlight}{HTML}{e5e5e5}
\tcbset{
  imgbox/.style={
    blankest,          % 내용만 남기는 가장 비어있는 스킨
    frame hidden,      % 프레임 감추기
    interior style={}, % 내부 채움 없음
    colback=white,     % (혹시 모를) 배경색 화이트
    colframe=white,    % 프레임색 화이트
    boxrule=0pt,       % 프레임 두께 0
    boxsep=0pt,        % 내용과 프레임 사이 여백 0
    sharp corners,     % 둥근 모서리 제거
    nobeforeafter      % 위아래 여백 제거
  }
}

\begin{document}
%
%Efficient Few-Shot Learning with Prompts from Knowledge Graphs
\title{EPLKG: Efficient Prompt Learning with Knowledge Graph}
\author{
Yongtaek Lim\textsuperscript{*}\inst{1} \and
Suho Kang\textsuperscript{*}\inst{2} \and
Yewon Kim\textsuperscript{*}\inst{3} \and
Dokyung Yoon\inst{3} \and
Kyungwoo Song\inst{2}
}

\institute{
DATUMO, Seoul, South Korea \and
Department of Statistics and Data Science, Yonsei University \and
Department of Artificial Intelligence, University of Seoul\\
\textsuperscript{*}Equal contribution.
}
%\author{First Author\inst{1}\orcidID{0000-1111-2222-3333} \and
%Second Author\inst{2,3}\orcidID{1111-2222-3333-4444} \and
%Third Author\inst{3}\orcidID{2222--3333-4444-5555}}
%\institute{Princeton University, Princeton NJ 08544, USA \and
%Springer Heidelberg, Tiergartenstr. 17, 69121 Heidelberg, Germany
\maketitle              % typeset the header of the contribution
\begin{abstract}
Large-scale pre-trained models such as CLIP excel in transferability and robust generalization across diverse datasets. However, adapting these models to new datasets or domains is computationally costly, especially in low-resource or few-shot settings, and existing prompt-learning methods often lack interpretability. We introduce Efficient Prompt Learning with Knowledge Graph (EPLKG), which uses a knowledge graph to curate diverse, interpretable prompts and, where KG coverage is limited, augments this bank with LLM-generated human-readable visual descriptions. \framework\ operates entirely on cached CLIP image and text embeddings and employs a lightweight Gumbel-Softmax module to select a single prompt per image–class pair, enabling low-memory, fast training. Across 11 benchmarks, EPLKG reduces per-image training time by up to 45\% and peak GPU memory by around 30--40\% compared to strong prompt-learning baselines, while keeping the average base--new harmonic-mean accuracy within 2 percentage points, thereby improving the efficiency of model adaptation without sacrificing competitive performance or interpretability.

\keywords{Multimodal learning \and Prompt learning \and Knowledge graph.}
\end{abstract}
\section{Introduction}
\label{sec:intro}
%\subsection{A Subsection Sample}
Large-scale pre-trained models have become essential for various tasks, showcasing remarkable transferability and competitive performance across diverse downstream applications, even with limited datasets. Their success has ignited extensive research in computer vision \cite{chen2020simple,grill2020bootstrap} and natural language processing \cite{devlin2018bert,brown2020language}. Recently, there has been a surge in efforts to expand large-scale pre-trained models into the realm of multimodal learning, capable of handling heterogeneous datasets encompassing both images and text. A notable multimodal model, Contrastive Language-Image Pre-training (CLIP) ~\cite{radford2021learning}, has shown significant performance improvements in various multimodal tasks, including retrieval ~\cite{so2022geodesic} and caption generation ~\cite{mokady2021clipcap}. CLIP stands out for its high transferability, excelling in zero-shot image classification by utilizing the text embedding module instead of a traditional classifier. This design allows CLIP to not only perform effective zero-shot classification but also to demonstrate competitive results in few-shot learning scenarios.
\begin{figure*}[t]\vspace{-0.5em}
    	\centering
    	\includegraphics[width=0.8\linewidth]
            {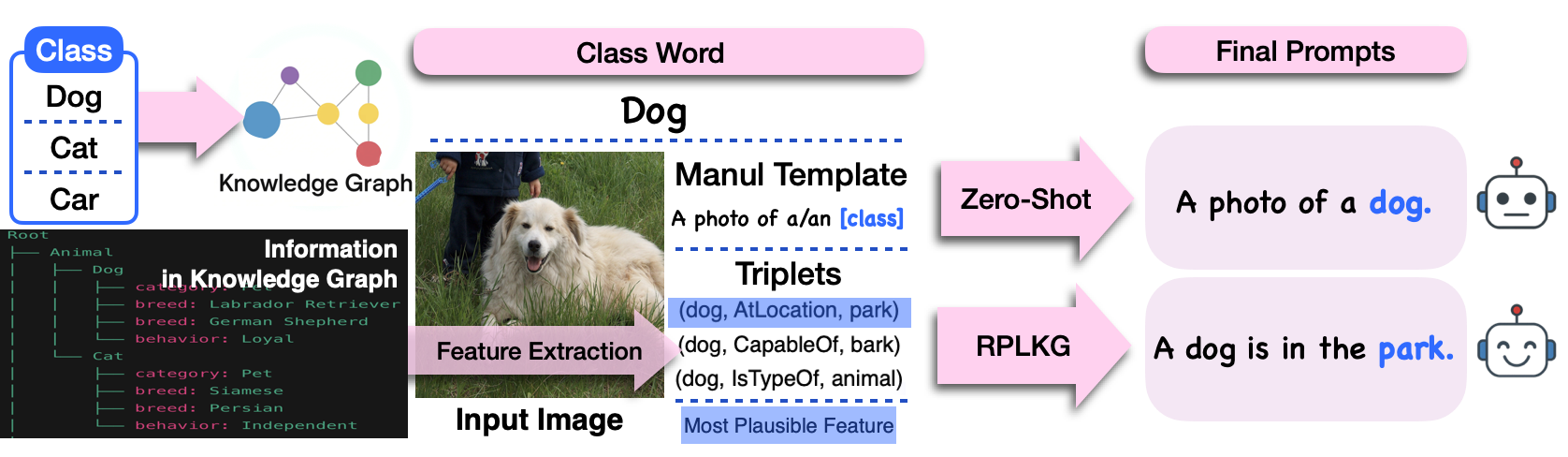}\vspace{-0.5em}
    \caption{
    The motivation behind \framework\ is to leverage a knowledge graph to extract semantically relevant information for each class, forming triplets that capture plausible attributes of class labels and generating interpretable prompts from these structured relations. The resulting prompt enhances both the model’s classification performance and interpretability, surpassing that of conventional zero-shot prompts.
    }\vspace{-1.5em}
    \label{fig:intro} 
\end{figure*}
However, adapting CLIP to new datasets by fine-tuning its parameters incurs high computational and memory costs. To mitigate this, several methods focus on training only a small subset of parameters, and prompt learning has emerged as a promising approach that introduces learnable parameters into the input layer to guide model behavior~\cite{zhou2022learning,zhou2022conditional}. This paper introduces \framework\, an efficient method for enhancing few-shot learning in large-scale pre-trained models. \framework\ uses a knowledge graph (KG) to generate text prompts, automatically selecting the optimal prompt for specific images. This approach addresses the challenge of creating large-scale prompt texts, which presents a combinatorial problem, through a Gumbel-Softmax-based prompt selection module \cite{jang2017categorical,maddison2017concrete}. \framework\ introduces three key enhancements over prior prompt learning approaches. First, it is computationally efficient, adding only 0.79M trainable parameters (about 0.5\% of CLIP) and avoiding backpropagation through the CLIP encoders. 
Second, it leverages human knowledge by converting KG triples into plain-text prompts, and third, it preserves interpretability by keeping prompts in natural language rather than continuous embeddings. As illustrated in Figure~\ref{fig:intro}, \framework\ extracts semantically relevant triples from the KG and aligns them with class labels to construct informative prompts.\newline

\renewcommand\labelitemi{$\bullet$}
The contributions of this work are summarized as follows:
\begin{itemize}
    \item We propose an optimal prompt selection mechanism that combines attention with Gumbel-Softmax on cached CLIP embeddings, adding only 0.79M parameters (about 0.5\% of CLIP).
    \item We design a simple framework to build interpretable plain-text prompts from a knowledge graph and augment them with LLM-generated visual descriptions when KG coverage is limited.
    \item We validate \framework\ on 11 benchmarks under few-shot, base--new, and domain generalization settings, reducing per-image training time (up to 45\%) and peak memory (around 30--40\%) while maintaining competitive accuracy.
\end{itemize}
%\item We introduce an optimal prompt selection mechanism using attention and Gumbel-Softmax with minimal learnable parameters atop the CLIP layer, reducing extra computational needs.
%\item We propose a framework to construct interpretable plain text prompts from KG with minimal rules, streamlining the conversion process.
%\item To rigorously evaluate \framework\, we conduct experiments on 11 datasets across multiple few-shot settings ($\{1, 2, 4, 8, 16\}$ shots), demonstrating its competitive generalization performance across domains.\vspace{-1.0em}

\begin{figure*}[t!]
	\centering
	\includegraphics[width=0.8\textwidth]
        {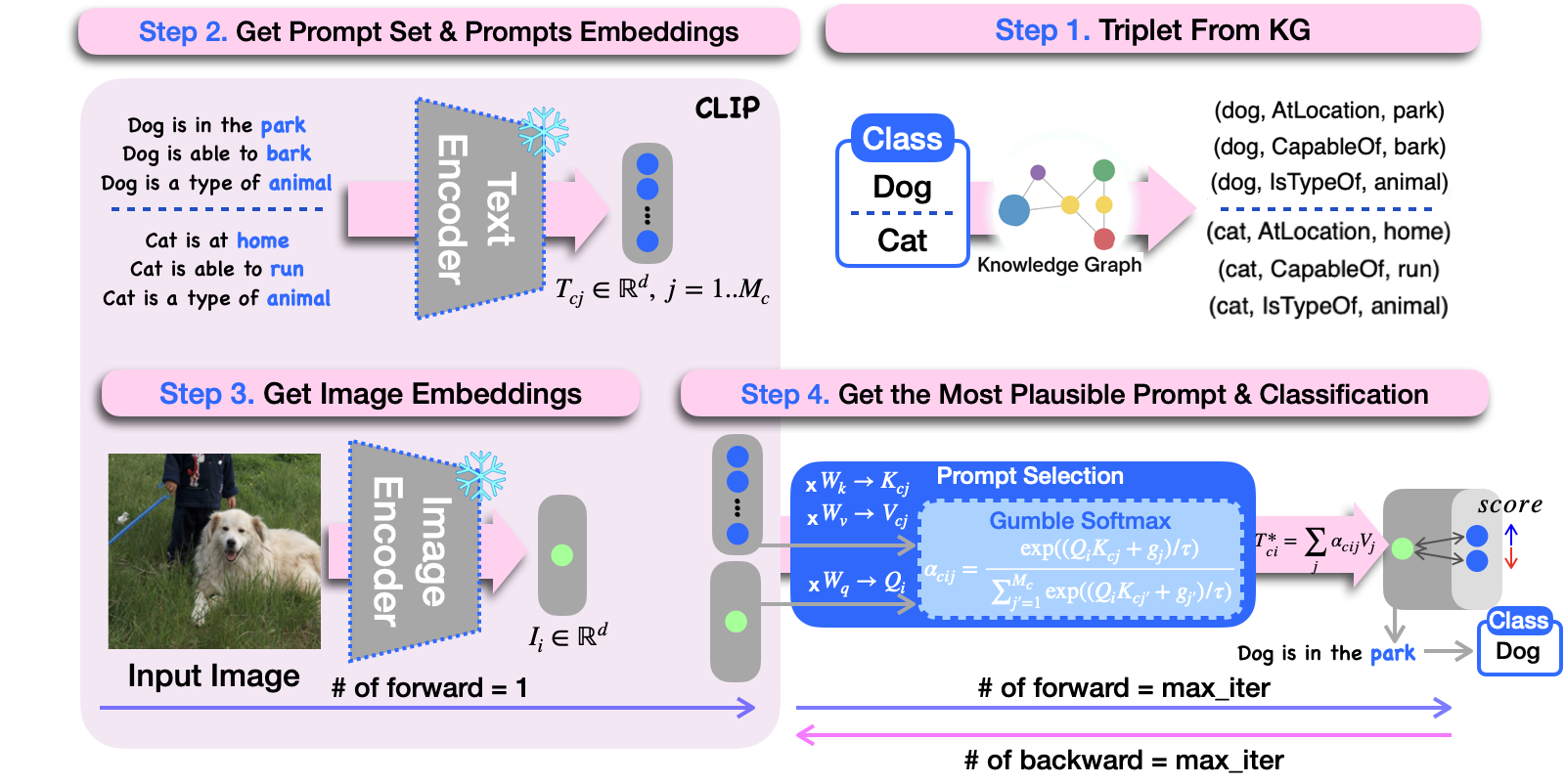}
	\vspace{-1.0em}
  \caption{Overview of \framework. For each class, KG triplets are converted into textual prompts and embedded with a frozen text encoder, while the input image is encoded with a frozen image encoder. \framework\ applies Gumbel-Softmax over CLIP image–text cosine similarities to select a compatible prompt for classification. Only the selection module is trained on cached embeddings, without backpropagation through the CLIP encoders, resulting in lower memory and compute cost and faster training.
}
  \vspace{-1.5em}
\label{fig:method_figure}
\end{figure*}

\section{Related Work}
\label{sec:related_work}
\subsection{Multimodal Learning} % 0.5page 이상 (영어로)
% CLIP, VirTEX, ... 
Recent research has focused on bridging the gap between natural language processing and computer vision by leveraging images and textual data, optimizing their alignment to develop reliable vision-language models (VLMs). To achieve this, efforts have been made to improve VLMs using bidirectional neural networks  \cite{wu2017cascade} and attention mechanisms \cite{li2017gla}. In particular, the field of Visual Question Answering (VQA) has highlighted the increasing importance of multimodal learning \cite{desta2018object}. Large-scale pre-trained models, such as VirTex \cite{desai2021virtex} and CLIP \cite{radford2021learning} have demonstrated broad applicability and substantial performance improvements. However, these models' zero-shot and few-shot learning capabilities have yet to reach their full potential \cite{wang2024benchmarkingzeroshotrobustnessmultimodal}. Consequently, substantial research efforts have been dedicated to addressing and improving vision-language alignment abilities, often through prompt learning \cite{gu2021ppt, zhou2022learning, zhou2022conditional, yao2023visuallanguageprompttuningknowledgeguided}. For instance, a more advanced approach involves utilizing LLM-generated descriptions that explicitly highlight discriminative features among visually similar classes, thereby advancing both performance and clarity \cite{lee2025enhancing}.

\subsection{Prompt Learning}
Prompt learning has become influential in NLP, effectively leveraging pre-trained knowledge of large language models (LLMs) through "fill-in-the-blank" style prompts, traditionally crafted manually. Due to the substantial impact of prompt templates on model performance, optimal prompt selection is essential. However, manual prompt creation is resource-intensive, driving the development of automated methods like Autoprompt \cite{autoprompt:emnlp20}. Furthermore, the advent of soft prompts allows mapping prompts to learnable embeddings, enabling end-to-end learning in a continuous space and diverging from traditional word-based formats \cite{li2021prefixtuning,lester2021power}. Extending beyond NLP, prompt learning has been applied to VLMs, significantly enhancing few-shot visual recognition and generalization. Innovations such as CoOp \cite{zhou2022learning} and CoCoOp \cite{zhou2022conditional} introduced continuous prompts to pre-trained models, while PLOT \cite{chen2022prompt} leveraged multiple prompts aligned with an image's local features for fine-grained tasks. KgCoOp \cite{yao2023visuallanguageprompttuningknowledgeguided} improves open-domain generalization via knowledge-guided prompting, but the interpretability of learned prompts (i.e., human-readable rationales) remains underexplored. Recent work uses GPT-3/4–generated visual descriptions to yield interpretable prompts and improve transfer to novel categories \cite{menon2023visual,maniparambil2023enhancingclipgpt4harnessing}. In contrast, our method, \framework\, retrieves class- and attribute-level prompts from a knowledge-graph–conditioned repository at inference time, providing human-interpretable prompts without fine-tuning the VLM and offering a lightweight alternative.

%vspace \cite{CHEN2025113118}, KG-Prompt \cite{zhang2024vspaceknowledgegraphbased} kg subsection으로

\subsection{Knowledge Graph }
KG has been widely used to inject structured prior knowledge into neural models. More recently, several works have explored using KG to guide prompt construction for large language models. vspace \cite{CHEN2025113118} extracts paths from a KG via reinforcement learning and turns them into prompts to enhance black-box LLMs such as GPT-3.5 \cite{brown2020language}. While effective, this design may introduce some hyperparameter sensitivity and additional compute. while KG-Prompt \cite{zhang2024knowgptknowledgegraphbased} presents an enhanced approach that leverages structured KG without requiring any modification to the architecture of LLMs. In contrast, \framework\ uses KGs in a vision–language setting with CLIP~\cite{radford2021learning}. We convert KG triples associated with each class into natural-language prompts and learn only a lightweight prompt-selection module on top of frozen CLIP embeddings, rather than backprop through the CLIP encoders. 
This design keeps the KG-induced prompts interpretable while enabling efficient adaptation with minimal additional parameters.

\section{Methodology}
\label{sec:method}
\subsection{Cached Embeddings}
To efficiently adapt large-scale pre-trained models, we decouple prompt learning from repeated forward–backward passes through CLIP. \framework\ operates on cached embeddings: As shown in Figure~\ref{fig:method_figure}, we precompute and store the text embeddings of all KG-derived prompts and the image embeddings of the training set using frozen CLIP encoders. During training, only a lightweight Gumbel-Softmax–based prompt selector and classification head are updated on top of these cached embeddings, while all CLIP parameters remain fixed. This design removes the need for backpropagation through the CLIP encoders and confines optimization to a small number of parameters, substantially reducing training time and peak memory usage compared with conventional prompt learning methods. Further efficiency analysis is given in Section~\ref{subsection:complexity}.

\subsection{Prompt with Knowledge Graph}
We leverage the KG ConceptNet, which facilitates simple, scalable prompt design and incorporation of human knowledge. ConceptNet provides a structured representation of entities and their relationships for prompt construction. Given a label identified as $entity_{i}$, a query to ConceptNet yields a triplet $(entity_{i}, relation_{ij},\\ entity_{j})$, such as (\textit{headphone}, \textit{UsedFor}, \textit{listening to music}). We then convert each triplet into a natural-language sentence, for example ``A headphone is used for listening to music.'' In this form, $entity_{j}$ represents a concept semantically associated with $entity_{i}$, and $relation_{ij}$ is realized as a relation-specific phrase (e.g., ``is used for'', ``is capable of'', ``is typically found in'') that describes how $entity_{i}$ and $entity_{j}$ are related. While ConceptNet is effective for generating semantically rich prompts on general-purpose datasets, it struggles with fine-grained or domain-specific labels such as those in FGVC-Aircraft or Stanford Cars. When a class is missing from ConceptNet or our prompt bank, we perform a one-time GPT-4.1 captioning pass on the few-shot training images. By guiding the LLMs to extract visual features rather than abstract descriptions, we enhance the model’s ability to focus on discriminative cues during inference. This method not only compensates for the limitations of symbolic resources in rigid domains but also improves the interpretability and effectiveness of the resulting prompts.

\subsection{Optimal Prompt Selection}
\begin{figure}[h]\vspace{-1.0em}
    \centering
    \begin{minipage}[bp]{0.5\textwidth}\normalsize
    In this section, we present our prompt optimization technique, which focuses on selecting the optimal prompt for each instance from a diverse set of interpretable text prompts. Traditional methods face a combinatorial challenge, with time complexity denoted as $M_{c}^{NC}$, where $N$ is the number of data instances, $C$ the number of classes, and $M_{c}$ the number of prompts for each class $c$. This approach is impractical for large-scale datasets due to its exponential time complexity. To address this, we
    \end{minipage}%
    \hfill  
    \begin{minipage}[bp]{0.5\textwidth}
    \begingroup
    {\footnotesize
    \begin{align}
    Q_{i}  &= I_{i} W_{q}, \nonumber \\
    K_{cj} &= T_{cj} W_{k}, \nonumber \\
    V_{cj} &= T_{cj} W_{v}, \label{eq:transform}
    \end{align}\vspace{-2.0em}
    }
    {
    \footnotesize
    \begin{align}
    u_{j} &\sim \text{Uniform}(0,1) \label{eq:gumbel_start} \\
    g_{j} &= -\log(-\log(u_{j})) \label{eq:gumbel_mid} \\
    \alpha_{cij} &= \frac{\exp((Q_{i}K_{cj}+g_{j})/\tau)}{\sum_{j'=1}^{M_{c}}\exp{((Q_{i}K_{cj'}+g_{j'})/\tau)}} \label{eq:gumbel}\\
    T_{ci}^{*} &= \sum_{j}\alpha_{cij} V_{cj} \label{eq:optimal_prompt}\\
    \widehat{c} &= \text{argmax}_{c'}I_{i} \cdot T_{ci}^{*} \label{eq:prediction}
    \end{align}
    }
    \endgroup
    \end{minipage}\vspace{-2.5em}
\end{figure}
\hspace{-1.5em}propose a Gumbel-Softmax-based method, \framework, which uses cached embeddings for images ($I$) and prompts ($T$) to optimize prompt selection efficiently. For each $i$-th image embedding $I_{i}$, we apply a linear transformation using weights $W_{q}$. Similarly, for each $j$-th prompt embedding of class $c$, $T_{cj}$, we apply transformations with $W_{k}$ and $W_{v}$. We then calculate the similarity between transformed image queries $Q_{i}$ and key vectors $K_{cj}$ for all prompts $j=1,...,M_{c}$, applying Gumbel-Softmax to normalize the similarity scores, as opposed to traditional Softmax, to select a single, optimal prompt. This choice is due to Gumbel-Softmax's ability to produce sparse output vectors, allowing for selecting a singular optimal prompt, which is both performance-enhancing and more interpretable. To implement the Gumbel-Softmax, we sample the random variable from the Gumbel (0,1) distribution by using inverse transformation, as shown in Equation \ref{eq:gumbel_start}-\ref{eq:gumbel_mid}. With the Gumbel random variable, we compute the softmax function by following \cite{jang2017categorical, maddison2017concrete}, as shown in Equation \ref{eq:gumbel}. Upon selecting the optimal prompt $T_{ci}^{}$ for the $i$-th image and $c$-th class, we compute the similarity between $I_{i}$ and $T_{ci}^{}$ for all classes $c=1,...,C$, choosing the class $\widehat{c}$ that maximizes this similarity. The model's parameters, $W{q}, W_{k}, W_{v}$, are optimized using cross-entropy loss between the ground-truth label $c^{*}$ and the predicted $\widehat{c}$. \framework's approach is characterized by its computational efficiency, requiring fewer resources than traditional prompt learning approaches.
\begin{table}[t!]
\caption{Comparison of accuracy and training time for prompt-based learning methods. \framework\ improves training speed while maintaining competitive accuracy; red text indicates the accuracy gap to the best baseline, and blue text indicates our relative training-time reduction (\%) compared to that baseline.}
\vspace{0.4em}
\centering
\resizebox{0.7\columnwidth}{!}{
\begin{tabular}{llcccr}
\toprule
Methods & Prompts & \multicolumn{3}{c}{Accuracy} & Training-time \\ 
\cmidrule(lr){3-5}
                 &                  & Base      & New       & H         &                        \\ 
\midrule
CLIP             & hand-crafted     & 69.34     & 74.22     & 71.70     & -                      \\
\midrule
CoOp             & textual          & \textbf{82.63}     & 67.99     & 74.60     & 6ms/image              \\
ProGrad          & textual          & 82.48     & 70.75     & 76.16     & 22ms/image             \\
CoCoOp            & textual+visual   & 80.47     & 71.69     & 75.83     & 160ms/image            \\
KgCoOp           & textual          & 80.73     & \textbf{73.60}     & \textbf{77.00}     & 6ms/image              \\
\textbf{\framework} & textual        & 81.18 (\textcolor{red}{-1.75\%})   & 71.37 (\textcolor{red}{-3.03\%})    & 75.96 (\textcolor{red}{-1.35\%})   & \textbf{3.3ms/image (\textcolor{blue}{-45\%})}           \\ 
\bottomrule
\end{tabular}
}\vspace{-1.0em}
\label{tab:time}
\end{table}

\begin{table}[t!]
\caption{We compare the peak memory (MB) of CoOp, CoCoOp, KgCoOp, and our model, \framework\, at training time. EPLKG shows the lowest peak memory usage across all batch sizes on both DTD and Oxford Pets datasets.}
\label{tab:peak_memory}
\centering
% ========== DTD (위) ==========
\resizebox{0.65\linewidth}{!}{%
\begin{tabular}{lcccc}
\multicolumn{5}{c}{\textbf{(a) Peak memory on DTD.}} \\[0.2em]  % ← 소제목 역할
\toprule
\multirow{2}{*}{Model} & \multicolumn{4}{c}{Batch Size} \\ \cmidrule(lr){2-5}
                       & 1 & 4 & 32 & 64 \\ \midrule
CoOp     & 988 & 988 & 1{,}005 & 1{,}025 \\
CoCoOp   & 988 & 3{,}101 & 22{,}943 & 45{,}615 \\
KgCoOp   & 1{,}038 & 1{,}040 & 1{,}056 & 1{,}075 \\ \midrule
\textbf{\framework}    
         & \textbf{625 (\textcolor{blue}{-36.74\%})} 
         & \textbf{625 (\textcolor{blue}{-36.74\%})} 
         & \textbf{632 (\textcolor{blue}{-37.11\%})} 
         & \textbf{651 (\textcolor{blue}{-36.49\%})} \\ 
\bottomrule
\end{tabular}
}%
\vspace{0.4em}
% ========== Oxford Pets (아래) ==========
\centering
\resizebox{0.65\linewidth}{!}{%
\begin{tabular}{lcccc}
\multicolumn{5}{c}{\textbf{(b) Peak memory on Oxford Pets.}} \\[0.2em]
\toprule
\multirow{2}{*}{Model} & \multicolumn{4}{c}{Batch Size} \\ \cmidrule(lr){2-5}
                       & 1 & 4 & 32 & 64 \\ \midrule
CoOp     & 988 & 988 & 1{,}005 & 1{,}025 \\
CoCoOp   & 988 & 3{,}101 & 22{,}943 & 45{,}615 \\
KgCoOp   & 913 & 915 & 924 & 929 \\ \midrule
\textbf{\framework}    
         & \textbf{625 (\textcolor{blue}{-31.54\%})} 
         & \textbf{625 (\textcolor{blue}{-31.69\%})} 
         & \textbf{632 (\textcolor{blue}{-31.60\%})} 
         & \textbf{651 (\textcolor{blue}{-29.92\%})} \\ 
\bottomrule
\end{tabular}
}
\end{table}

\section{Results}
\subsection{Experimental Settings}
For few-shot classification, following prior work \cite{menon2022visual, roth2023waffling, lee2025enhancing}, we use ViT-B/16 and evaluate 11 datasets with 1, 2, 4, 8, and 16 shots per class. For domain generalization, we train ViT-B/16 and ViT-B/32 on ImageNet with 16 shots per class and evaluate on multiple ImageNet variants. For base-to-new generalization, following \cite{zhou2022conditional, zhou2022learning}, we train CoOp, CoCoOp, KgCoOp, and \framework\ on base classes with 16 shots and test on all classes. We search over $\alpha \in \{0.1, 0.3, 0.5, 0.7, 1.0\}$, weight decay $\in \{3\text{e-3}, 1\text{e-2}, 5\text{e-2}, 0.1\}$, Gumbel-Softmax temperature $\in \{0.001, 0.01, 0.1\}$, and dropout $\in \{0.1, 0.3, 0.5, 0.7\}$, and use the prompt template “A photo of [CLASS]” for zero-shot CLIP.

\subsection{Time and Memory Efficiency} \label{subsection:complexity}
In this section, we evaluate the peak memory usage and training time of \framework\ compared to CoOp and CoCoOp, KgCoOp on the DTD and Oxford Pets datasets, aiming to achieve optimal performance. These metrics are assessed under a 16-shot setting with a batch size 64. The training times for the baselines and \framework\ are reported in Table \ref{tab:time}. \framework\ achieves competitive accuracy while reducing training time by ~45\% compared to the strong baseline KgCoOp, with $<$4\% drop in performance. Moreover, Tables \ref{tab:peak_memory}(a) and \ref{tab:peak_memory}(b) detail our model's memory efficiency, demonstrating that \framework\ requires less memory across different batch sizes compared to the baselines. In essence, \framework\ achieves comparable results to existing prompt learning methods but with markedly lower time and memory requirements. 

\begin{figure}[t!]
\centering
\scalebox{0.8}{
\begin{tcbraster}[
    raster columns=3,
    raster column skip=-0.5mm,  % 가로 간격
    raster row skip=0mm,     % 세로 간격
    raster left skip=0pt,
    raster right skip=0pt
]
  % 한 장짜리 PDF면 width만 주면 됨
  \tcbincludegraphics[imgbox,graphics options={width=\linewidth}]{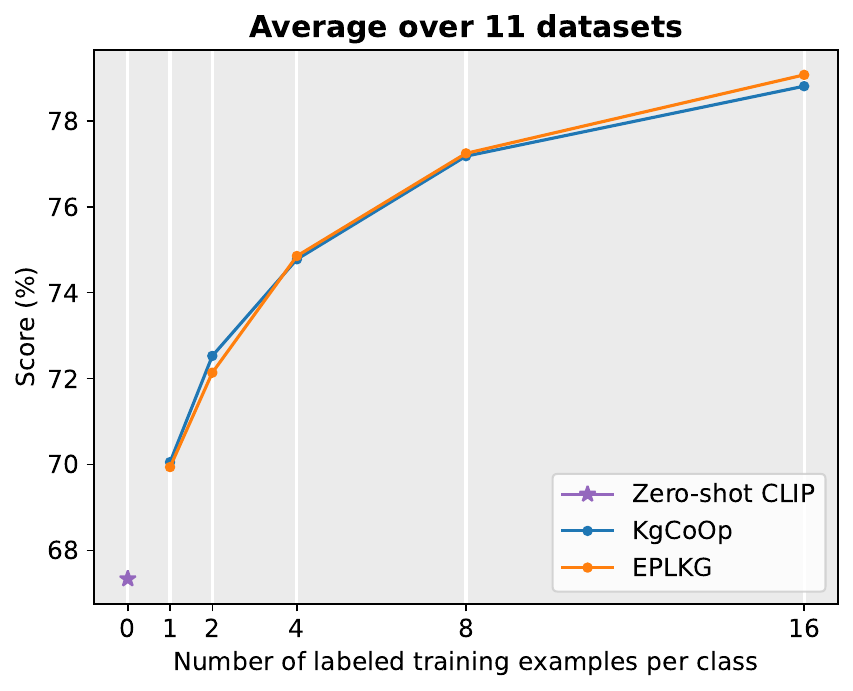}
  \tcbincludegraphics[imgbox,graphics options={width=\linewidth}]{figures/main_curves/Imagenet.pdf}
  \tcbincludegraphics[imgbox,graphics options={width=\linewidth}]{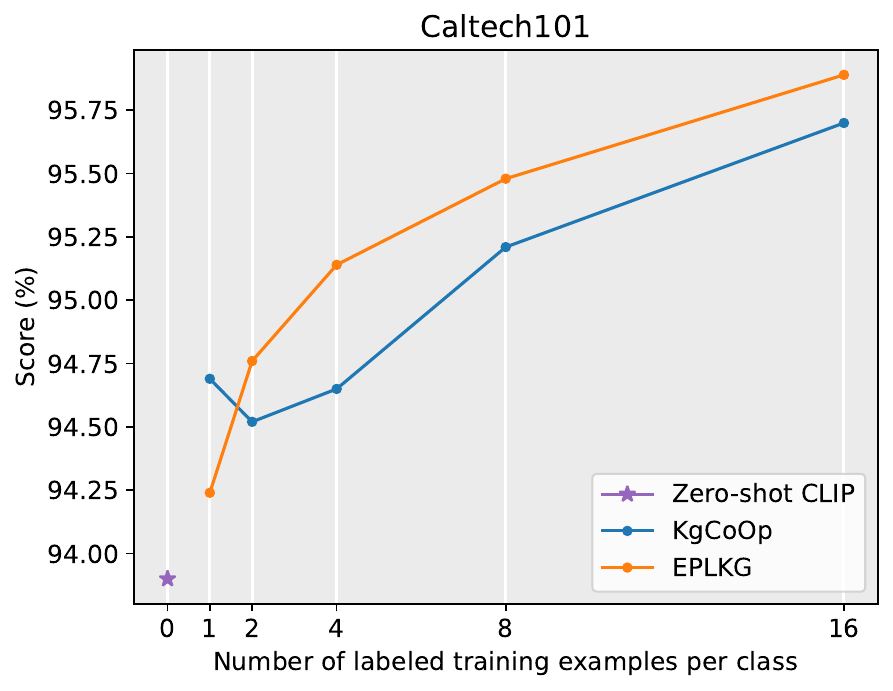}
  \tcbincludegraphics[imgbox,graphics options={width=\linewidth}]{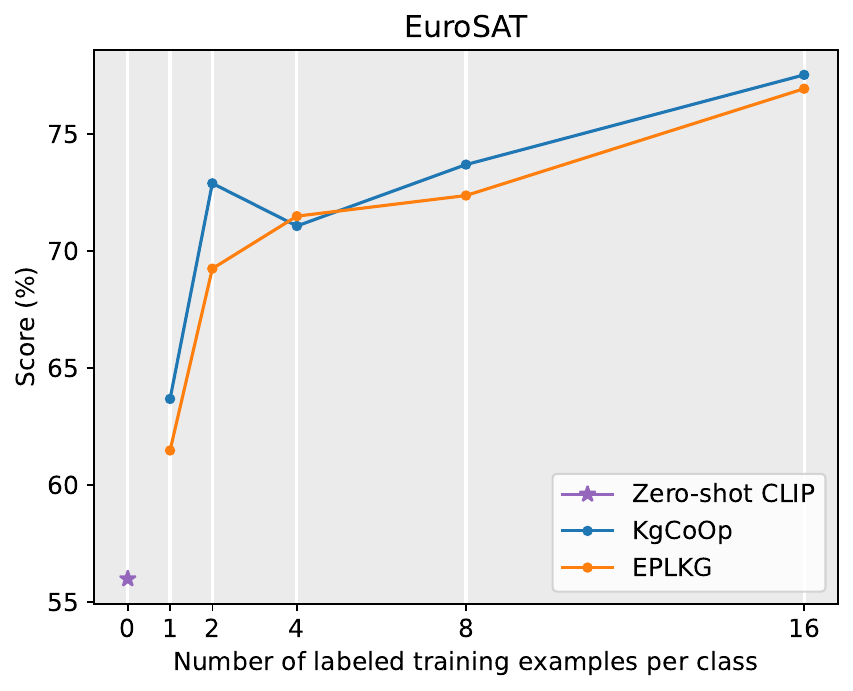}
  \tcbincludegraphics[imgbox,graphics options={width=\linewidth}]{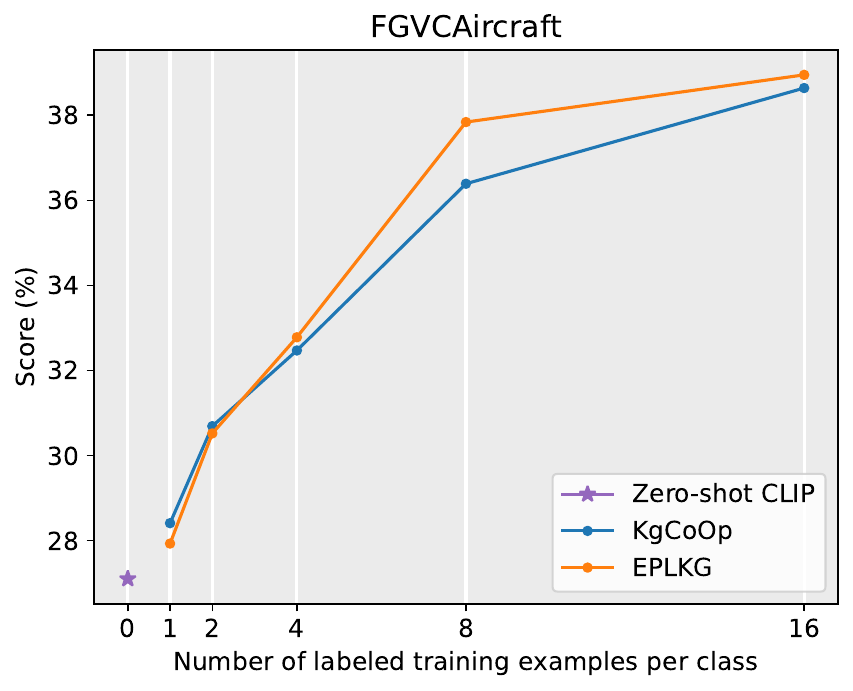}
  \tcbincludegraphics[imgbox,graphics options={width=\linewidth}]{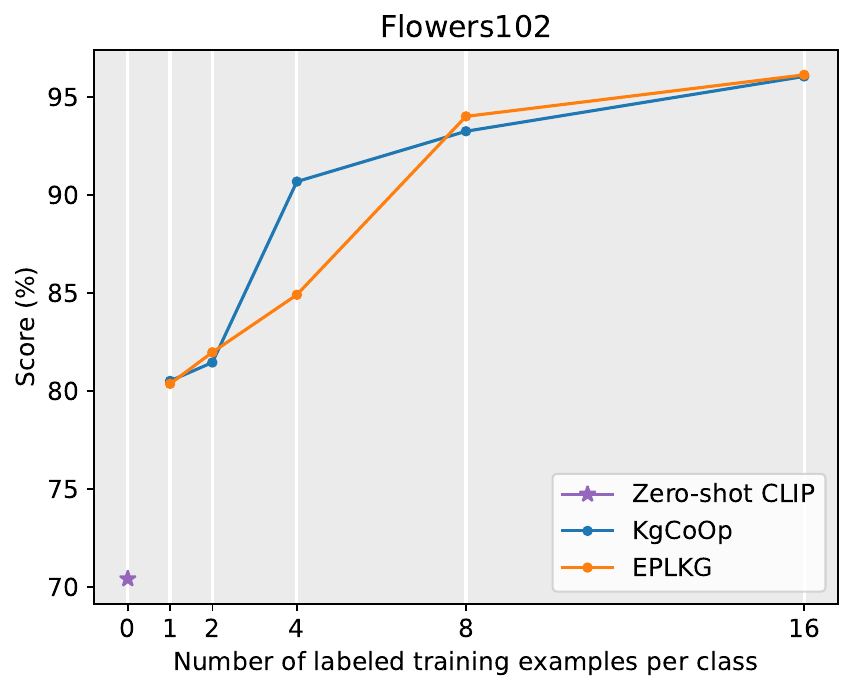}
  \tcbincludegraphics[imgbox,graphics options={width=\linewidth}]{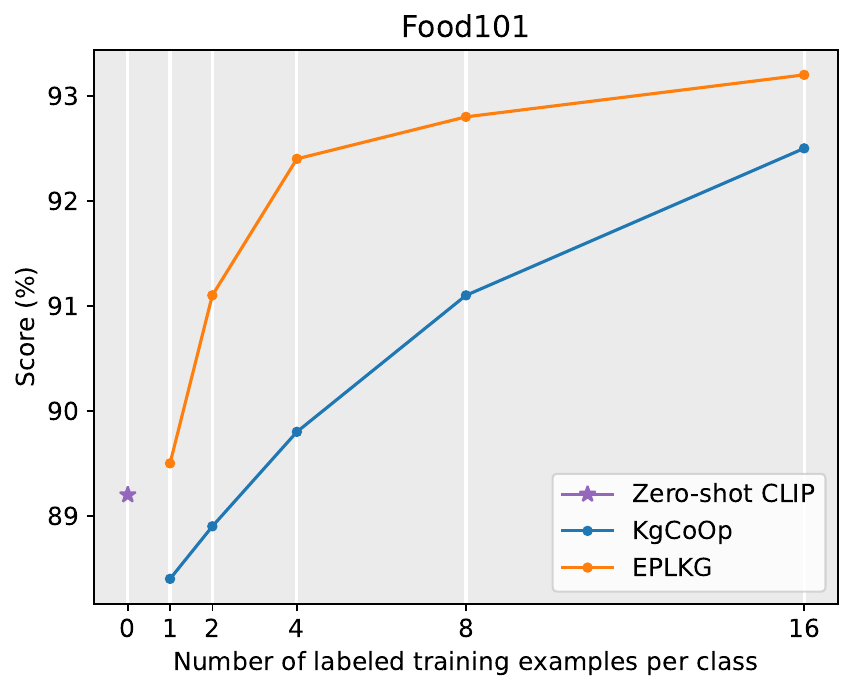}
  \tcbincludegraphics[imgbox,graphics options={width=\linewidth}]{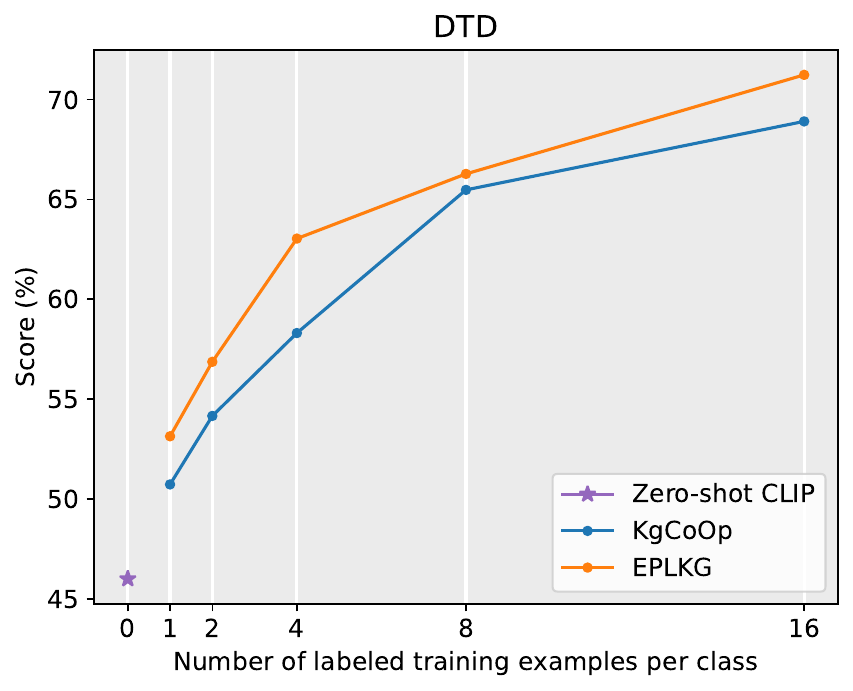}
  \tcbincludegraphics[imgbox,graphics options={width=\linewidth}]{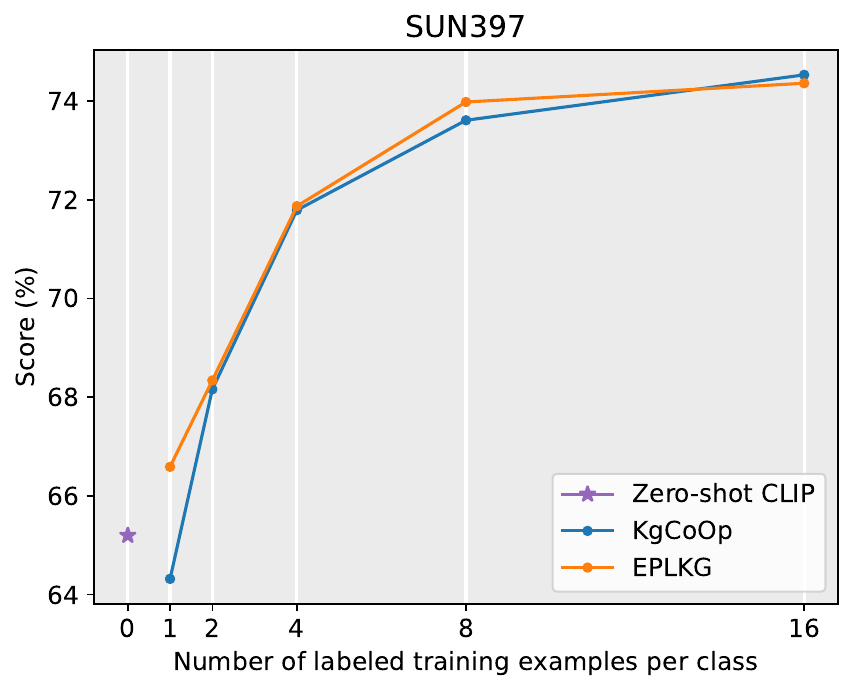}
  \tcbincludegraphics[imgbox,graphics options={width=\linewidth}]{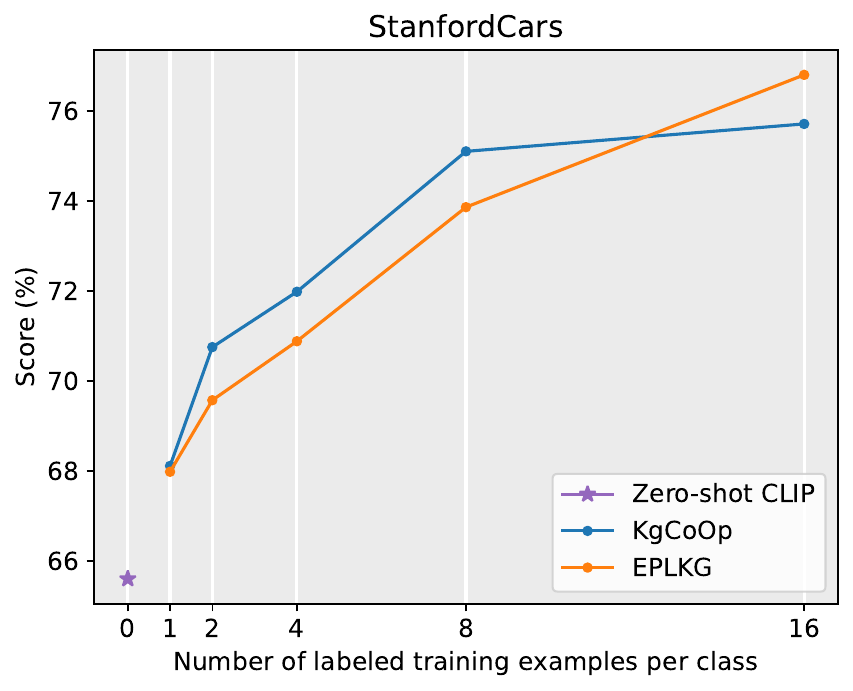}
  \tcbincludegraphics[imgbox,graphics options={width=\linewidth}]{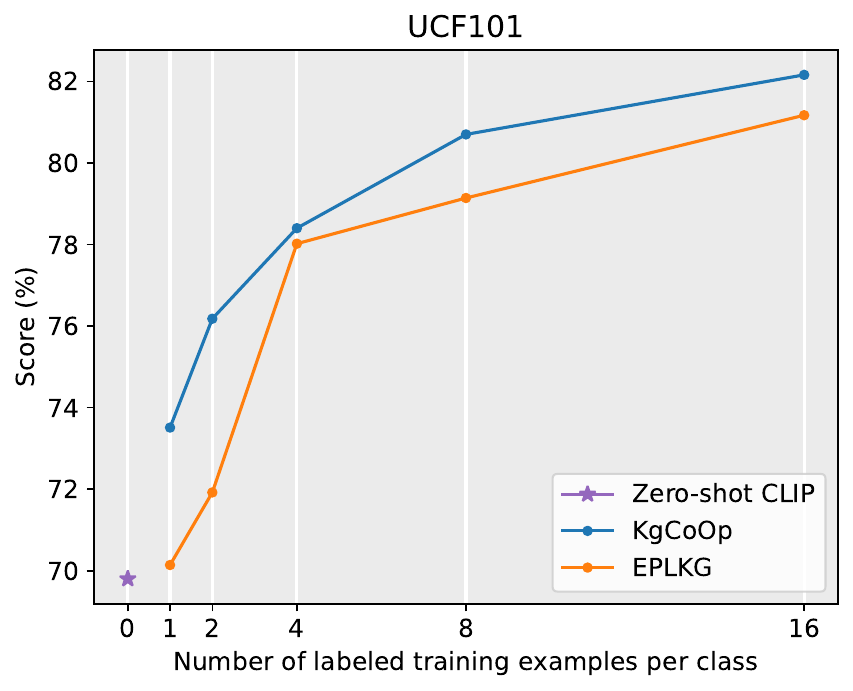}
  \tcbincludegraphics[imgbox,graphics options={width=\linewidth}]{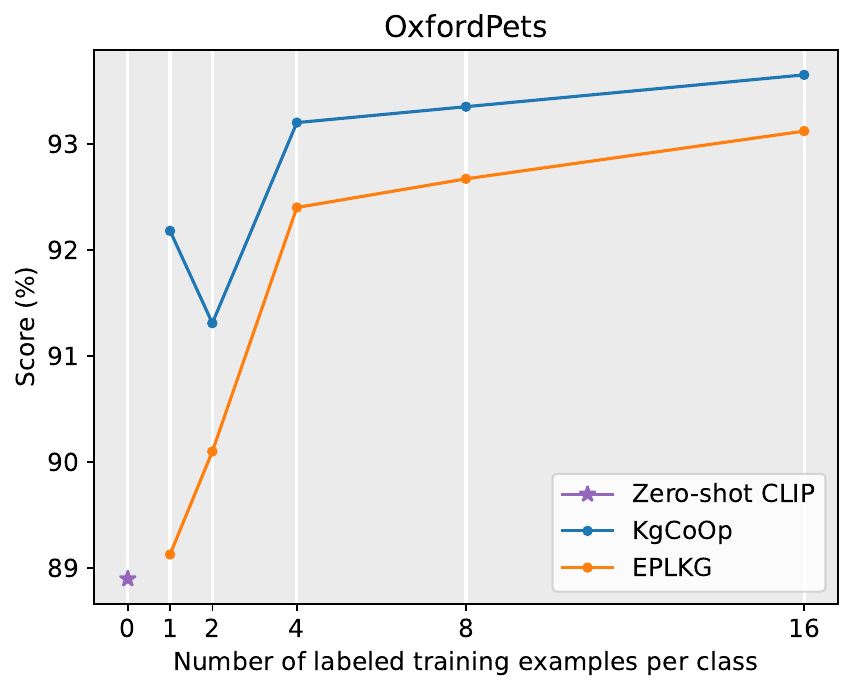}
\end{tcbraster}
}
\caption{The first subplot shows the average performance over 11 datasets; the others show \framework\ and baseline performance per dataset for $k \in \{1,2,4,8,16\}$ shots. 
}
\label{fig:fewshot_performance}
\end{figure}

\subsection{Few-shot Classification}
We conduct few-shot classification experiments on 11 datasets with k in {1, 2, 4, 8, 16}-shots. As shown in Figure \ref{fig:fewshot_performance}, our model, \framework\, is compared against Zero-shot CLIP and a strong baseline, KgCoOp. The results clearly demonstrate that \framework\ outperforms or matches on most datasets both baselines across almost all datasets and shot counts. Notably, the performance gap is significant even with a single training example (1-shot), and this advantage is maintained as the number of shots increases, highlighting our method's effectiveness and scalability in few-shot learning scenarios.

\begin{table}[t!]
    \caption{Base-to-new performance on 11 datasets. Following Zhou et al.~\cite{zhou2022conditional}, we train CoOp, CoCoOp, KgCoOp, and \framework\ on base classes with 16 shots and test on all classes. Despite its lower training-time and memory footprint, \framework\ attains the second-best average base--new harmonic mean across all 11 datasets, competitive with the strongest baseline. Best and second-best results are shown in \textbf{bold} and \underline{underlined}, respectively; H denotes the harmonic mean.}
    \tabstyle{2pt}
    \scriptsize
    \begin{subtable}{0.3\textwidth}
    \centering
    \caption{average of 11 datasets.}\vspace{-0.5em}
    \begin{tabular}{l cc|c}
    \toprule
    & Base & New & H \\
    \midrule
    CLIP & 69.34 & \textbf{74.22} & 71.70 \\
    CoOp & \textbf{82.69} & 63.22 & 71.66 \\
    CoCoOp & 80.47 & 71.69 & 75.83 \\
    KgCoOp & 80.73 & \underline{73.60} & \textbf{77.00} \\
    \rowcolor{tabhighlight}
    \framework\ & \underline{81.18} & 71.37 & \underline{75.96} \\
    \bottomrule
    \end{tabular}
    \end{subtable}
    ~\vspace{-0.5em}
    \begin{subtable}{0.3\textwidth}
    \centering
    \caption{ImageNet}\vspace{-0.5em}
    \begin{tabular}{l cc|c}
    \toprule
    & Base & New & H \\
    \midrule
    CLIP & 72.43 & 68.14 & 70.22 \\
    CoOp & \textbf{76.47} & 67.88 & 71.92\\
    CoCoOp & 75.98 & \textbf{70.43} & \textbf{73.10} \\
    KgCoOp & 75.83 & \underline{69.96} & \underline{72.78} \\
    \rowcolor{tabhighlight}
    \framework\ & \underline{76.30} & 66.47 & 71.05 \\
    \bottomrule
    \end{tabular}
    \end{subtable}
    ~\vspace{-0.5em}
    \begin{subtable}{0.3\textwidth}
    \centering
    \caption{Caltech101.}\vspace{-0.5em}
    \begin{tabular}{l cc|c}
    \toprule
    & Base & New & H \\
    \midrule
    CLIP & 96.84 & \underline{94.00} & 95.40 \\
    CoOp & \underline{98.00} & 89.81 & 93.73 \\
    CoCoOp & 97.96 & 93.81 & \underline{95.84} \\
    KgCoOp & 97.72 & \textbf{94.39} & \textbf{96.03} \\
    \rowcolor{tabhighlight}
    \framework\ & \textbf{98.04} & 93.56 & 95.75 \\
    \bottomrule
    \end{tabular}
    \end{subtable}
    ~\vspace{-0.5em}
    \begin{subtable}{.3\textwidth}
    \centering
    \caption{OxfordPets.}\vspace{-0.5em}
    \begin{tabular}{l cc|c}
    \toprule
    & Base & New & H \\
    \midrule
    CLIP & 91.17 & 97.26 & 94.12 \\
    CoOp & 93.67 & 95.29 & 94.47 \\
    CoCoOp & \textbf{95.20} & \underline{97.69} & \textbf{96.43} \\
    KgCoOp & \underline{94.65} & \textbf{97.76} & \underline{96.18} \\
    \rowcolor{tabhighlight}
    \framework\ & 93.83 & 96.33 & 95.06 \\
    \bottomrule
    \end{tabular}
    \end{subtable}
    \vspace{1em}
    ~\vspace{-0.5em}
    \begin{subtable}{.3\textwidth}
    \centering
    \caption{StanfordCars.}\vspace{-0.5em}
    \begin{tabular}{l cc|c}
    \toprule
    & Base & New & H \\
    \midrule
    CLIP & 63.37 & \underline{74.89} & 68.65 \\
    CoOp & \textbf{78.12} & 60.40 & 68.13 \\
    CoCoOp & 70.49 & 73.59 & 72.01 \\
    KgCoOp & 71.76 & \textbf{75.04} & \textbf{73.36} \\
    \rowcolor{tabhighlight}
    \framework\ & \underline{73.67} & 71.58 & \underline{72.61} \\
    \bottomrule
    \end{tabular}
    \end{subtable}
    ~\vspace{-0.5em}
    \begin{subtable}{.3\textwidth}
    \centering
    \caption{Flowers102.}\vspace{-0.5em}
    \begin{tabular}{l cc|c}
    \toprule
    & Base & New & H \\
    \midrule
    CLIP & 72.08 & \textbf{77.80} & 74.83 \\
    CoOp & \textbf{97.60} & 59.67 & 74.06 \\
    CoCoOp & 94.87 & 71.75 & \underline{81.71} \\
    KgCoOp & 95.00 & \underline{74.73} & \textbf{83.65} \\
    \rowcolor{tabhighlight}
    \framework\ & \underline{96.52} & 70.04 & 81.18 \\
    \bottomrule
    \end{tabular}
    \end{subtable}
    ~\vspace{-0.5em}
    \begin{subtable}{.3\textwidth}
    \centering
    \caption{Food101.}\vspace{-0.5em}
    \begin{tabular}{l cc|c}
    \toprule
    & Base & New & H \\
    \midrule
    CLIP & 90.10 & 91.22 & 90.66 \\
    CoOp & 88.33 & 82.26 & 85.19 \\
    CoCoOp & \textbf{90.70} & \underline{91.29} & \underline{90.99} \\
    KgCoOp & \underline{90.50} & \textbf{91.70} & \textbf{91.09} \\
    \rowcolor{tabhighlight}
    \framework\ & 89.68 & 88.54 & 89.10 \\
    \bottomrule
    \end{tabular}
    \end{subtable}
    \vspace{1em}
    ~\vspace{-0.5em}
    \begin{subtable}{.3\textwidth}
    \centering
    \caption{FGVCAircraft.}\vspace{-0.5em}
    \begin{tabular}{l cc|c}
    \toprule
    & Base & New & H \\
    \midrule
    CLIP & 27.19 & \textbf{36.29} & 31.09 \\
    CoOp & \textbf{40.44} & 22.30 & 28.75 \\
    CoCoOp & 33.41 & 23.71 & 27.74 \\
    KgCoOp & \underline{36.21} & \underline{33.55} & \textbf{34.83} \\
    \rowcolor{tabhighlight}
    \framework\ & 35.55 & 31.07 & \underline{33.16} \\
    \bottomrule
    \end{tabular}
    \end{subtable}
    ~\vspace{-0.5em}
    \begin{subtable}{.3\textwidth}
    \centering
    \caption{SUN397.}\vspace{-0.5em}
    \begin{tabular}{l cc|c}
    \toprule
    & Base & New & H \\
    \midrule
    CLIP & 69.36 & 75.35 & 72.23 \\
    CoOp & \underline{80.60} & 65.89 & 72.51 \\
    CoCoOp & 79.74 & \textbf{76.86} & \underline{78.27} \\
    KgCoOp & 80.29 & \underline{76.53} & \textbf{78.36} \\
    \rowcolor{tabhighlight}
    \framework\ & \textbf{80.86} & 74.26 & 77.42 \\
    \bottomrule
    \end{tabular}
    \end{subtable}
    ~\vspace{-0.5em}
    \begin{subtable}{.3\textwidth}
    \centering
    \caption{DTD.}\vspace{-0.5em}
    \begin{tabular}{l cc|c}
    \toprule
    & Base & New & H \\
    \midrule
    CLIP & 53.24 & \textbf{59.90} & 56.37 \\
    CoOp & \underline{79.44} & 41.18 & 54.24 \\
    CoCoOp & 77.01 & \underline{56.00} & \underline{64.85} \\
    KgCoOp & 77.55 & 54.99 & 64.35 \\
    \rowcolor{tabhighlight}
    \framework\ & \textbf{81.13} & 54.95 & \textbf{65.52} \\
    \bottomrule
    \end{tabular}
    \end{subtable}
    ~\vspace{-0.5em}
    \begin{subtable}{.3\textwidth}
    \centering
    \caption{EuroSAT.}\vspace{-0.5em}
    \begin{tabular}{l cc|c}
    \toprule
    & Base & New & H \\
    \midrule
    CLIP & 56.48 & 64.05 & 60.03 \\
    CoOp & \textbf{92.19} & 54.74 & 68.69 \\
    CoCoOp & \underline{87.49} & 60.04 & 71.21 \\
    KgCoOp & 85.64 & \underline{64.34} & \underline{73.48} \\
    \rowcolor{tabhighlight}
    \framework\ & 84.18 & \textbf{65.30} & \textbf{73.55} \\
    \bottomrule
    \end{tabular}
    \end{subtable}
    ~\vspace{-0.5em}
    \begin{subtable}{.3\textwidth}
    \centering
    \caption{UCF101.}\vspace{-0.5em}
    \begin{tabular}{l cc|c}
    \toprule
    & Base & New & H \\
    \midrule
    CLIP & 70.53 & \textbf{77.50} & 73.85 \\
    CoOp & \textbf{84.69} & 56.05 & 67.46 \\
    CoCoOp & 82.33 & 73.45 & 77.64 \\
    KgCoOp & 82.89 & \underline{76.67} & \textbf{79.65} \\
    \rowcolor{tabhighlight}
    \framework\ & \underline{83.25} & 72.99 & \underline{77.78} \\
    \bottomrule
    \end{tabular}
    \end{subtable}
\label{tab:results_base2new}
\end{table}

\subsection{New Class Generalization}
Table \ref{tab:results_base2new} presents the results for the base-to-new generalization setting, which evaluates how well models generalize to new classes after being trained on base classes. Despite using significantly less peak memory and requiring shorter training times than CoOp, CoCoOp, and KgCoOp, our model, \framework\, consistently achieves competitive performance across all 11 datasets. Notably, \framework\ attains a competitive overall harmonic mean, indicating strong generalization. These results show that \framework\ is efficient and scalable while matching or surpassing more resource-intensive baselines. The strong generalization capability of \framework\ is attributed to its robust prompt selection mechanism. Unlike CoOp and CoCoOp, which optimize continuous prompt vectors from scratch, \framework\ first generates a diverse set of candidate prompts derived from triplets in a structured knowledge base. It then leverages a few training examples to select the optimal prompt, which in turn enhances its generalization performance without the need for resource-intensive, end-to-end parameter optimization.

\begin{table}[t!]
\centering
\caption{Domain generalization performance on ImageNet variants. Models are trained on ImageNet and evaluated in a zero-shot setting on multiple ImageNet variants. \framework\ consistently outperforms the baselines across all datasets and backbone networks.}
\label{table:robustness_various_archs}
\vspace{0.4em}
\resizebox{0.55\linewidth}{!}{
\begin{tabular}{l ccccc}
\toprule
 & Source & \multicolumn{4}{c}{Target} \\
\cmidrule(lr){2-2} \cmidrule(lr){3-6}
Method & ImageNet & -V2 & -Sketch & -A & -R \\
\midrule
\textbf{ViT-B/32} \\
Zeroshot-CLIP      & 62.06 & 54.76 & 40.83 & 29.83 & 65.93 \\
Linear Probe CLIP  & 59.58 & 49.73 & 28.06 & 19.67 & 47.20 \\
CLIP + \framework\ & \textbf{66.11} & \textbf{57.07} & \textbf{41.11} & \textbf{30.19} & \textbf{66.19} \\
\midrule
\textbf{ViT-B/16} \\
Zeroshot-CLIP      & 66.73 & 60.85 & 46.16 & 47.80 & 73.99 \\
Linear Probe CLIP  & 65.85 & 56.26 & 34.77 & 35.68 & 58.43 \\
CLIP + \framework\ & \textbf{71.74} & \textbf{64.01} & \textbf{46.67} & \textbf{48.49} & \textbf{74.94} \\
\bottomrule
\end{tabular}
}
\end{table}

\subsection{Domain Generalization}
We conducted domain generalization experiments to assess the robustness of our model under distribution shifts. Using the evaluation protocol of \cite{zhou2022learning} and the datasets listed in Table~\ref{table:robustness_various_archs}, we found that \framework\ consistently outperforms Zeroshot-CLIP across all ImageNet variants and backbone architectures. These results highlight the benefit of incorporating structured knowledge into prompt learning. Moreover, unlike Linear Probe CLIP—which is trained only on the source domain and degrades significantly under shift—our approach maintains stable performance. By integrating external semantic knowledge from a KG, \framework\ enables the model to learn more robust and invariant representations. This improvement suggests that external knowledge sources, when properly aligned with visual representations through prompt optimization, can provide meaningful inductive biases. These biases help bridge the domain gap and preserve semantic consistency, thereby enhancing robustness in real-world scenarios where data distributions are often non-stationary. Our findings highlight the potential of \framework\ not just as a tool for improving few-shot classification, but also as a promising method for domain-agnostic transfer learning.

\begin{figure}[t!]
  % 모든 서브캡션 위쪽 간격을 살짝만 두기
  \captionsetup{skip=2pt}

  \centering
  % ① ImageNet ---------------------------------------------------
  \begin{subfigure}[b]{\linewidth}
    \centering
    \includegraphics[width=.75\linewidth]{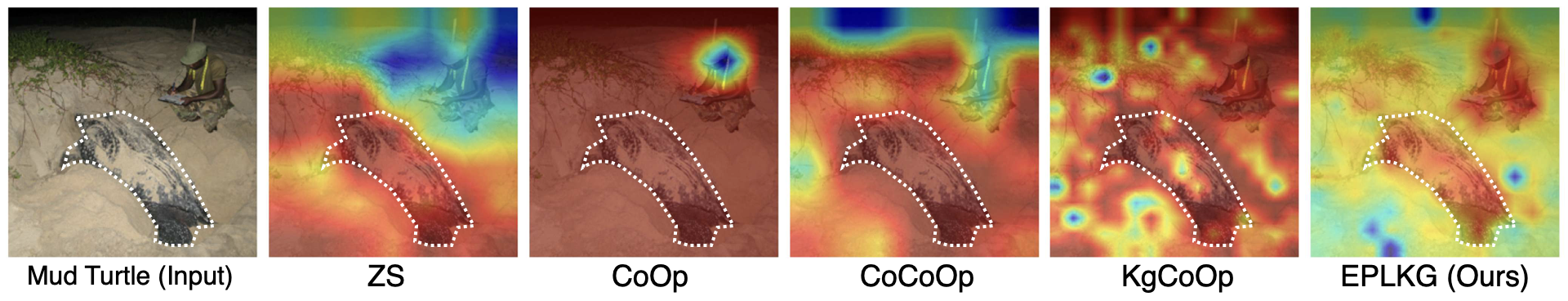}
    \vspace{-0.5em}
    \caption{ImageNet – Mud Turtle. Chosen Prompt: “A mud turtle has a low, rounded shell.”}
    \label{fig:gradcam_imagenet_row}
  \end{subfigure}\par\vspace{4pt}

  % ② Oxford Pets -----------------------------------------------
  \begin{subfigure}[b]{\linewidth}
    \centering
    \includegraphics[width=.75\linewidth]{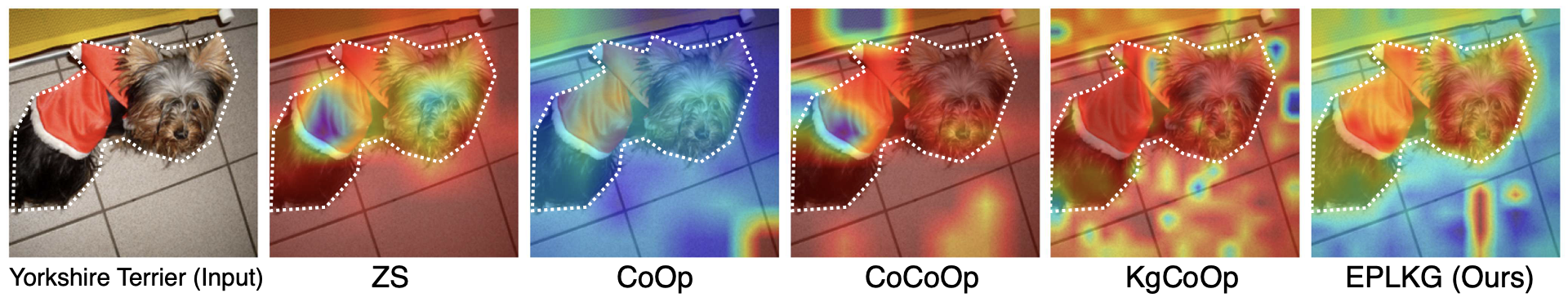}
    \vspace{-0.5em}
    \caption{Oxford Pets – Yorkshire Terrier. Chosen Prompt: “A Yorkshire Terrier has long, silky hair.”}
    \label{fig:gradcam_pets_row}
  \end{subfigure}\par\vspace{4pt}

  % ③ Food-101 --------------------------------------------------
  \begin{subfigure}[b]{\linewidth}
    \centering
    \includegraphics[width=0.75\linewidth]{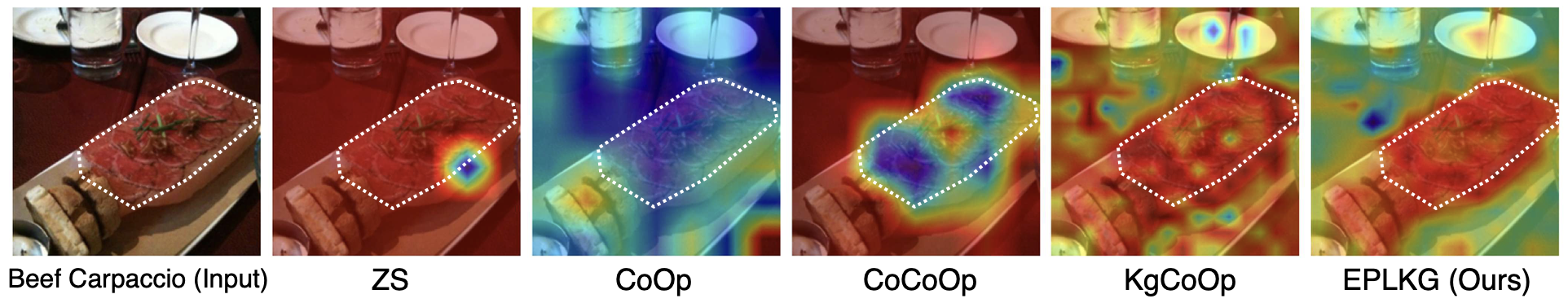}
    \vspace{-0.5em}
    \caption{Food-101 – Beef Carpaccio. Chosen Prompt: “Beef carpaccio shows thin slices of raw red meat.”}
    \label{fig:gradcam_food_row}
  \end{subfigure}\par\vspace{4pt}

  % ④ Flowers-102 ----------------------------------------------
  \begin{subfigure}[b]{\linewidth}
    \centering
    \includegraphics[width=.75\linewidth]{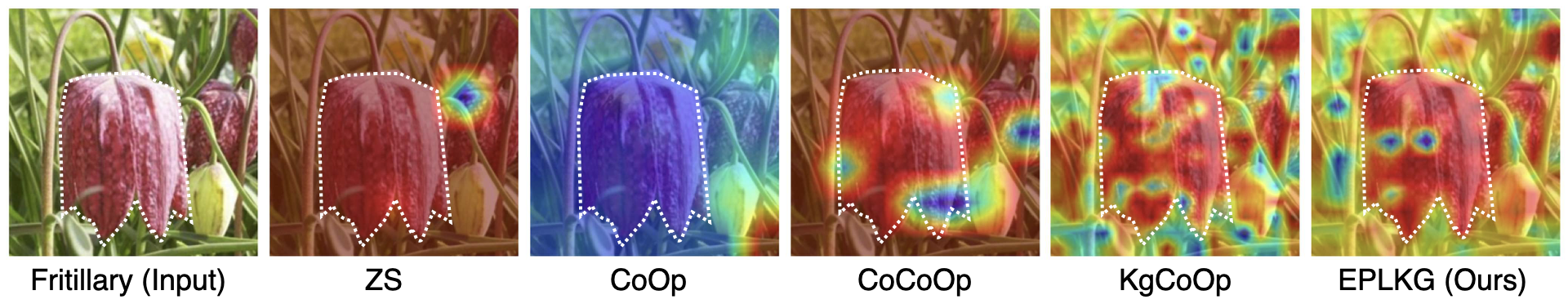}
    \vspace{-0.5em}
    \caption{Flowers-102 – Fritillary. Chosen Prompt: “A fritillary has a drooping, bell-shaped flower.”}
    \label{fig:gradcam_flowers_row}
  \end{subfigure}

  % ─── 메인(공통) 캡션 ───
  \caption{%
Each row presents Grad-CAM visualizations for zero-shot (ZS), CoOp, CoCoOp, KgCoOp and \framework\ (ours). Red regions indicate higher Grad-CAM relevance scores for the target class. In cases where the baseline models fail to make correct predictions, \framework\ succeeds by selecting an optimal prompt from the KG.
  }
  \label{fig:gradcam_total}
\end{figure}
\subsection{Grad-CAM Analysis} \label{subsection:gradcam_analysis}
As illustrated in Figure~\ref{fig:gradcam_total}, baseline methods such as ZS, CoOp, CoCoOp and KGCoOp frequently fail to capture the correct semantics or attend to task-relevant visual regions. In contrast, our proposed method, \framework\, exhibits more semantically grounded and interpretable attention patterns. Across various dataset examples, \framework\ consistently selects prompts that closely align with the target class attributes. For instance, in the case of the \emph{Mud Turtle}, while other methods tend to focus on irrelevant background regions or overly broad areas surrounding the fish, \framework\ precisely attends to the fish itself. The Grad-CAM reveals that it focuses on discriminative features in a more concentrated and human-aligned manner. These results provide compelling evidence of \framework's superior interpretability and robustness in prompt-based learning.

\section{Conclusion}
We presented \framework, an efficient prompt learning framework that builds interpretable KG- and LLM-derived prompts and selects them with a lightweight Gumbel-Softmax module on cached CLIP embeddings. Across 11 datasets and few-shot, base--new, and domain generalization settings, \framework\ achieves competitive accuracy while reducing training time by up to 45\% and peak memory by 30--40\%, and keeping the base--new harmonic mean within a few percentage points of the best baseline; Grad-CAM analyses show that it focuses on semantically meaningful regions. While effective in our experiments, \framework\ may be sensitive to KG coverage and quality and to the stochastic nature of Gumbel-Softmax, especially in domains with sparse or noisy knowledge. Future work includes leveraging domain-specific KGs, designing stable or uncertainty-aware selection, and applying \framework\ to continual, online, and federated learning scenarios where parameter-efficient, interpretable adaptation is crucial.
\bibliographystyle{splncs04}
\bibliography{main.bib}

@inproceedings{radford2021learning,
  title={Learning transferable visual models from natural language supervision},
  author={Radford, Alec and Kim, Jong Wook and Hallacy, Chris and Ramesh, Aditya and Goh, Gabriel and Agarwal, Sandhini and Sastry, Girish and Askell, Amanda and Mishkin, Pamela and Clark, Jack and others},
  booktitle={International Conference on Machine Learning},
  pages={8748--8763},
  year={2021},
  organization={PMLR}
}

@article{so2022geodesic,
  title={Geodesic Multi-Modal Mixup for Robust Fine-Tuning},
  author={So, Junhyuk and Oh, Changdae and Lim, Yongtaek and Byun, Hoyoon and Shin, Minchul and Song, Kyungwoo},
  journal={arXiv preprint arXiv:2203.03897},
  year={2022}
}

@article{mokady2021clipcap,
  title={Clipcap: Clip prefix for image captioning},
  author={Mokady, Ron and Hertz, Amir and Bermano, Amit H},
  journal={arXiv preprint arXiv:2111.09734},
  year={2021}
}

@article{zhou2022learning,
  title={Learning to prompt for vision-language models},
  author={Zhou, Kaiyang and Yang, Jingkang and Loy, Chen Change and Liu, Ziwei},
  journal={International Journal of Computer Vision},
  volume={130},
  number={9},
  pages={2337--2348},
  year={2022},
  publisher={Springer}
}

@inproceedings{zhou2022conditional,
  title={Conditional prompt learning for vision-language models},
  author={Zhou, Kaiyang and Yang, Jingkang and Loy, Chen Change and Liu, Ziwei},
  booktitle={Proceedings of the IEEE/CVF Conference on Computer Vision and Pattern Recognition},
  pages={16816--16825},
  year={2022}
}

@misc{yao2023visuallanguageprompttuningknowledgeguided,
      title={Visual-Language Prompt Tuning with Knowledge-guided Context Optimization}, 
      author={Hantao Yao and Rui Zhang and Changsheng Xu},
      year={2023},
      eprint={2303.13283},
      archivePrefix={arXiv},
      primaryClass={cs.CV},
      url={https://arxiv.org/abs/2303.13283}, 
}

@inproceedings{desai2021virtex,
  title={Virtex: Learning visual representations from textual annotations},
  author={Desai, Karan and Johnson, Justin},
  booktitle={Proceedings of the IEEE/CVF conference on computer vision and pattern recognition},
  pages={11162--11173},
  year={2021}
}

@article{brown2020language,
  title={Language models are few-shot learners},
  author={Brown, Tom and Mann, Benjamin and Ryder, Nick and Subbiah, Melanie and Kaplan, Jared D and Dhariwal, Prafulla and Neelakantan, Arvind and Shyam, Pranav and Sastry, Girish and Askell, Amanda and others},
  journal={Advances in neural information processing systems},
  volume={33},
  pages={1877--1901},
  year={2020}
}

@article{grill2020bootstrap,
  title={Bootstrap your own latent-a new approach to self-supervised learning},
  author={Grill, Jean-Bastien and Strub, Florian and Altch{\'e}, Florent and Tallec, Corentin and Richemond, Pierre and Buchatskaya, Elena and Doersch, Carl and Avila Pires, Bernardo and Guo, Zhaohan and Gheshlaghi Azar, Mohammad and others},
  journal={Advances in neural information processing systems},
  volume={33},
  pages={21271--21284},
  year={2020}
}

@inproceedings{chen2020simple,
  title={A simple framework for contrastive learning of visual representations},
  author={Chen, Ting and Kornblith, Simon and Norouzi, Mohammad and Hinton, Geoffrey},
  booktitle={International conference on machine learning},
  pages={1597--1607},
  year={2020},
  organization={PMLR}
}

@inproceedings{jang2017categorical,
  title={Categorical Reparametrization with Gumble-Softmax},
  author={Jang, Eric and Gu, Shixiang and Poole, Ben},
  booktitle={International Conference on Learning Representations (ICLR 2017)},
  year={2017},
  organization={OpenReview. net}
}

@inproceedings{maddison2017concrete,
  title={The concrete distribution: A continuous relaxation of discrete random variables},
  author={Maddison, C and Mnih, A and Teh, Y},
  booktitle={Proceedings of the international conference on learning Representations},
  year={2017},
  organization={International Conference on Learning Representations}
}

@article{gu2021ppt,
  title={Ppt: Pre-trained prompt tuning for few-shot learning},
  author={Gu, Yuxian and Han, Xu and Liu, Zhiyuan and Huang, Minlie},
  journal={arXiv preprint arXiv:2109.04332},
  year={2021}
}

@article{devlin2018bert,
  title={Bert: Pre-training of deep bidirectional transformers for language understanding},
  author={Devlin, Jacob and Chang, Ming-Wei and Lee, Kenton and Toutanova, Kristina},
  journal={arXiv preprint arXiv:1810.04805},
  year={2018}
}

@article{wu2017cascade,
  title={Cascade recurrent neural network for image caption generation},
  author={Wu, Jie and Hu, Haifeng},
  journal={Electronics Letters},
  volume={53},
  number={25},
  pages={1642--1643},
  year={2017},
  publisher={Wiley Online Library}
}

@article{li2017gla,
  title={GLA: Global--local attention for image description},
  author={Li, Linghui and Tang, Sheng and Zhang, Yongdong and Deng, Lixi and Tian, Qi},
  journal={IEEE Transactions on Multimedia},
  volume={20},
  number={3},
  pages={726--737},
  year={2017},
  publisher={IEEE}
}

@inproceedings{desta2018object,
  title={Object-based reasoning in VQA},
  author={Desta, Mikyas T and Chen, Larry and Kornuta, Tomasz},
  booktitle={2018 IEEE Winter Conference on Applications of Computer Vision (WACV)},
  pages={1814--1823},
  year={2018},
  organization={IEEE}
}

@inproceedings{
menon2023visual,
title={Visual Classification via Description from Large Language Models},
author={Sachit Menon and Carl Vondrick},
booktitle={The Eleventh International Conference on Learning Representations },
year={2023},
url={https://openreview.net/forum?id=jlAjNL8z5cs}
}

@inproceedings{autoprompt:emnlp20,
  author = {Taylor Shin and Yasaman Razeghi and Robert L. Logan IV and Eric Wallace and Sameer Singh},
  title = { {AutoPrompt}: Eliciting Knowledge from Language Models with Automatically Generated Prompts },
  booktitle = {Empirical Methods in Natural Language Processing (EMNLP)},
  year = {2020}
}

@misc{li2021prefixtuning,
      title={Prefix-Tuning: Optimizing Continuous Prompts for Generation}, 
      author={Xiang Lisa Li and Percy Liang},
      year={2021},
      eprint={2101.00190},
      archivePrefix={arXiv},
      primaryClass={cs.CL}
}

@misc{
chen2022prompt,
title={Prompt Learning with Optimal Transport for Vision-Language Models},
author={Guangyi Chen and Weiran Yao and Xiangchen Song and Xinyue Li and Yongming Rao and Kun Zhang},
year={2022},
url={https://openreview.net/forum?id=b9APFSTylGT}
}

@article{lester2021power,
  title={The power of scale for parameter-efficient prompt tuning},
  author={Lester, Brian and Al-Rfou, Rami and Constant, Noah},
  journal={arXiv preprint arXiv:2104.08691},
  year={2021}
}

@article{menon2022visual,
  title={Visual classification via description from large language models},
  author={Menon, Sachit and Vondrick, Carl},
  journal={arXiv preprint arXiv:2210.07183},
  year={2022}
}

@inproceedings{roth2023waffling,
  title={Waffling around for performance: Visual classification with random words and broad concepts},
  author={Roth, Karsten and Kim, Jae Myung and Koepke, A and Vinyals, Oriol and Schmid, Cordelia and Akata, Zeynep},
  booktitle={Proceedings of the IEEE/CVF International Conference on Computer Vision},
  pages={15746--15757},
  year={2023}
}

@inproceedings{lee2025enhancing,
  title={Enhancing Visual Classification Using Comparative Descriptors},
  author={Lee, Hankyeol and Seo, Gawon and Choi, Wonseok and Jung, Geunyoung and Song, Kyungwoo and Jung, Jiyoung},
  booktitle={2025 IEEE/CVF Winter Conference on Applications of Computer Vision (WACV)},
  pages={5274--5283},
  year={2025},
  organization={IEEE}
}

@misc{wang2024benchmarkingzeroshotrobustnessmultimodal,
      title={Benchmarking Zero-Shot Robustness of Multimodal Foundation Models: A Pilot Study}, 
      author={Chenguang Wang and Ruoxi Jia and Xin Liu and Dawn Song},
      year={2024},
      eprint={2403.10499},
      archivePrefix={arXiv},
      primaryClass={cs.LG},
      url={https://arxiv.org/abs/2403.10499}, 
}

@misc{zhang2024knowgptknowledgegraphbased,
      title={KnowGPT: Knowledge Graph based Prompting for Large Language Models}, 
      author={Qinggang Zhang and Junnan Dong and Hao Chen and Daochen Zha and Zailiang Yu and Xiao Huang},
      year={2024},
      eprint={2312.06185},
      archivePrefix={arXiv},
      primaryClass={cs.CL},
      url={https://arxiv.org/abs/2312.06185}, 
}

@article{CHEN2025113118,
title = {KG-prompt: Interpretable knowledge graph prompt for pre-trained language models},
journal = {Knowledge-Based Systems},
volume = {311},
pages = {113118},
year = {2025},
issn = {0950-7051},
doi = {https://doi.org/10.1016/j.knosys.2025.113118},
url = {https://www.sciencedirect.com/science/article/pii/S0950705125001650},
author = {Liyi Chen and Jie Liu and Yutai Duan and Runze Wang}
}

@misc{maniparambil2023enhancingclipgpt4harnessing,
      title={Enhancing CLIP with GPT-4: Harnessing Visual Descriptions as Prompts}, 
      author={Mayug Maniparambil and Chris Vorster and Derek Molloy and Noel Murphy and Kevin McGuinness and Noel E. O'Connor},
      year={2023},
      eprint={2307.11661},
      archivePrefix={arXiv},
      primaryClass={cs.CV},
      url={https://arxiv.org/abs/2307.11661}, 
}
%Author, F.: Article title. Journal \textbf{2}(5), 99--110 (2016)

%\end{thebibliography}
\end{document}